\documentclass{article} 
\usepackage{iclr2026_conference,times}

\usepackage{amsmath,amsfonts,bm}

\def\eqref#1{equation~\ref{#1}}

\def\1{\bm{1}}

\DeclareMathAlphabet{\mathsfit}{\encodingdefault}{\sfdefault}{m}{sl}
\SetMathAlphabet{\mathsfit}{bold}{\encodingdefault}{\sfdefault}{bx}{n}

\usepackage{hyperref}
\usepackage{url}
\usepackage{booktabs}
\usepackage{multirow}
\usepackage{graphicx}
\usepackage{makecell}

\usepackage{graphicx}
\usepackage{subcaption}
\usepackage{setspace}

\title{NeuroTTT: Bridging Pretrain–Downstream Task Misalignment in EEG Foundation Models via Test-Time Training}

\author{Suli~Wang\textsuperscript{\rm 1}, 
Yangshen~Deng\textsuperscript{\rm 2}, 
Zhenghua~Bao\textsuperscript{\rm 1}, 
Xinyu~Zhan\textsuperscript{\rm 1}, 
Yiqun~Duan \textsuperscript{\rm 3}\thanks{Corresponding author.} \\
\textsuperscript{\rm 1}Technical University of Darmstadt, Germany\\
\textsuperscript{\rm 2}University of Edinburgh, UK\\
\textsuperscript{\rm 3}University of Technology Sydney, Australia\\
\texttt{\href{mailto:wsuli615@gmail.com}{wsuli615@gmail.com}, \href{mailto:yiqun.duan-1@uts.edu.au}{yiqun.duan-1@uts.edu.au}.} \\
}

\iclrfinalcopy 
\begin{document}

\maketitle

\begin{abstract}

Large-scale foundation models for EEG signals offer a promising path to generalizable brain–computer interface (BCI) applications, but they often suffer from misalignment between pretraining objectives and downstream tasks, as well as significant fine-tuning and test-time distribution shifts. We introduce NeuroTTT, a two-stage alignment strategy that bridges the gap between generic pretraining and task-specific EEG decoding tasks. First, we perform a domain-specific self-supervised fine-tuning paradigm that augments the foundation model with task-relevant self-supervised objectives, aligning latent representations to important spectral, spatial, and temporal EEG features without requiring additional labeled data. Second, we incorporate test-time training (TTT) at inference, we apply (i) self-supervised test-time training on individual unlabeled test samples and (ii) prediction entropy minimization (Tent), which updates only normalization statistics to continually calibrate the model to each new input on the fly. Our approach, which, to our knowledge, is the first to unify domain-tuned self-supervision with test-time training in large-scale EEG foundation models, yields substantially improved robustness and accuracy across diverse BCI tasks (imagined speech, stress detection, motor imagery). Using CBraMod and LaBraM as backbones, our method pushes their performance to a markedly higher level. Results on three diverse tasks demonstrate that the proposed alignment strategy achieves state-of-the-art performance, outperforming conventional fine-tuning and adaptation methods. Our code is available at~\url{https://github.com/wsl2000/NeuroTTT}.

\end{abstract}

\section{Introduction}

Electroencephalography (EEG) is a non-invasive technique for measuring brain electrical activity and underpins a variety of brain-computer interface (BCI) applications including but not limited to imagined speech decoding~\citep{proix2022imagined}, mental stress detection~\citep{badr2024review}, emotion recognition~\citep{li2022dynamic} and motor imagery classification~\citep{altaheri2023deep}. Early EEG decoding approaches heavily relied on handcrafted features and traditional machine learning~\citep{ramoser2000optimal,ang2008filter}. In recent years, deep learning models have achieved superior performance by learning directly from raw EEG signals in an end-to-end fashion~\citep{craik2019deep,al2021deep}. However, most of deep learning models employ supervised learning methods tailored for specific tasks or datasets, and lack generalization ability. Inspired by the success of foundation models in natural language processing (NLP) and computer vision (CV)~\citep{devlin2019bert,liu2024deepseek}, researchers begin developing large-scale EEG foundation models – pretrained foundation models intended to serve as general feature extractors for diverse EEG tasks~\citep{lai2025simple}. Most of these works leverage massive EEG datasets in generative-based self-supervised pretraining and finetuned on specific downstream datasets~\citep{yang2023biot,duan2023dewave}. These models have shown promising performance in capturing generic EEG representations and addressing issues like limited data and heterogeneous channel configurations.

\begin{figure}[!t]
    \centering
    \includegraphics[width=1.0\linewidth]{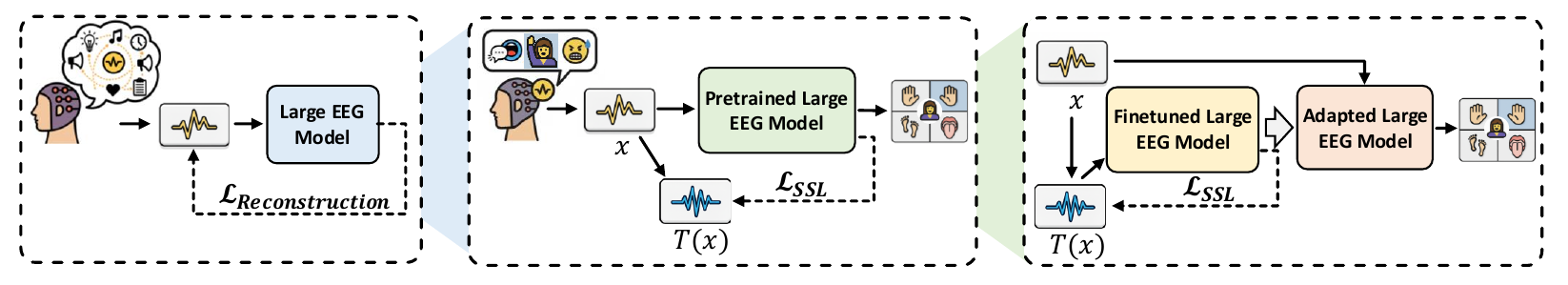}
    \caption{\textbf{NeuroTTT overview.} EEG foundation backbones are pre-trained on large, task-agnostic datasets, whereas fine-tuning targets a specific downstream task (e.g., motor imagery). Thus large EEG models often suffer from a mismatch between pretraining objectives and downstream requirements. We address this in two stages: Stage I—the backbone feeds a supervised task head together with lightweight self-supervised (SSL) heads to align representations toward task-relevant EEG features. Stage II—for each test sample, we apply test-time training with the SSL objective, or entropy minimization that updates only batch-normalization statistics, further mitigating residual mismatch.}
    \label{fig:placeholder}
\end{figure}

However, fine-tuning strategies for EEG foundation models remain relatively underexplored. A fundamental yet often overlooked issue is the misalignment between pre-training objectives/data and downstream EEG task requirements: generic self-supervised objectives (e.g., masked signal reconstruction or neural code prediction~\citep{he2022masked,wang2024cbramod,jiang2024large}) may fail to emphasize the task-critical spectral, spatial, or temporal features needed for downstream domain specific applications, leaving downstream decoders under-served without targeted alignment during adapting. Compounding this, domain shift is pervasive in EEG: inter- and intra-subject variability~\citep{saha2020intra}, session effects, and device/protocol differences routinely degrade cross-subject generalization~\citep{melnik2017systems}. Aligning the distribution of a target subject with that of the source subjects has long been a thorny challenge, motivating transfer learning and domain adaptation techniques for BCIs~\citep{zhao2020deep,lan2018domain,duan2023domain}.  Addressing both issues—\textbf{pretrain–downstream misalignment and distribution shift}—is therefore essential for robust EEG decoding. Simple adaptation strategies like training a shallow classifier (e.g. a linear or MLP head) on top of the pretrained foundation model often yield suboptimal results.  In practice, we observe that the features from these foundation models are not sufficiently aligned to task-specific patterns, making downstream adaptation challenging.

We hypothesize that introducing domain-specific self-supervised objectives during fine-tuning can help bridging the gap between pretraining and target tasks. By augmenting the foundation model with lightweight, task-relevant self-supervised learning (SSL) tasks, we can encourage the model to learn features that are more congruent with the downstream task’s discriminative information. Crucially, these SSL tasks are plug-and-play and do not require additional labeled data, we exploit unlabeled structure in EEG signals (such as frequency content or sensor topology) to shape the representation. This prepares a better feature space for final classification and yields improved performance with minimal overhead. Our SSL fine-tuning strategy is flexible: new self-supervised tasks can be designed as needed for different EEG contexts, leveraging known neuroscientific priors (e.g. known frequency bands or brain region asymmetries).

Beyond enhancing the feature space alignment, we also address model adaptation at inference time. EEG signals are notoriously variable across recording sessions and subjects\citep{corsi2007within}; thus even a well-finetuned model may face distribution shifts when deployed on new subjects or conditions. To tackle this, we employ Test-Time Training (TTT) techniques that adjust the model on-the-fly for each test sample or batch~\citep{sun2019test,sun2020test}. 
This test-time training with self-supervision effectively calibrates the model to the incoming data, improving robustness to shifts without requiring any labeled data or extensive retraining. Importantly, the adaptation is lightweight and privacy-friendly, all computations occur on the test sample itself with no need to share or store data from the source domain.
We further incorporate a complementary test-time strategy: entropy minimization of the model’s predictions (exploiting the Tent method for fully test-time training)~\citep{wang2020tent}. By minimizing the prediction entropy on unlabeled test inputs, we encourage the model’s outputs to be more confident and cluster test features toward the learned decision boundaries. This approach updates only a minimal subset of parameters (e.g. batch normalization statistics) during inference, providing a stable yet effective adjustment. Notably, we find that entropy-based TTA can outperform full-parameter self-supervised TTT in challenging cross-subject scenarios, likely because it avoids overfitting by tuning only normalization layers while leaving the backbone intact.

In summary, we propose a novel two-stage alignment approach for EEG foundation models, our main contributions can be summarized as follows:

\begin{enumerate}
    \item We propose NeuroTTT, a two-stage fine-tuning paradigm for EEG foundation models that first uses domain-specific self-supervised tasks to relieve distribution shift and pretrain–downstream misalignment. This approach is lightweight and general, adding no additional deployment cost while significantly aligning the representation to task-specific features.
    \item We propose to apply test-time self-supervised training for brain signal models, enabling on-the-fly calibration to individual subjects and trials without any labeled data or source model modifications.
    \item We explore test-time entropy minimization in EEG contexts and demonstrate its potential to improve generalization while offering a stable and privacy-preserving adaptation mechanism.
    \item We evaluate the performance of NeuroTTT on 3 different downstream tasks and demonstrate state-of-the-art results on representative EEG foundation models, suggesting that our approach is a practical step toward more task-aligned, generalizable brain foundation models.
\end{enumerate}



\section{Related Work}
\textbf{Large Brain Models:} The concept of foundation models in EEG is nascent, inspired by the breakthroughs of large language models and vision transformers. Early efforts like BENDR and EEG2Vec explored self-supervised pre-training on EEG~\citep{kostas2021bendr,bethge2022eeg2vec}, but were limited in scale and generality. Recent advances have produced truly large EEG models trained on extensive and diverse datasets. CBraMod (Criss-Cross Brain Model) is a transformer-based foundation model that separately models spatial and temporal dependencies of EEG through a criss-cross attention mechanism~\citep{wang2024cbramod}. Pre-trained via masked EEG patch reconstruction on a massive corpus, CBraMod achieved state-of-the-art results across numerous BCI tasks, demonstrating the potential of a single model to generalize widely. LaBraM (Large Brain Model) introduced a unified EEG representation learned from about 2,500 hours of data from public datasets~\citep{jiang2024large}. It uses a vector-quantized spectral tokenizer and masked token prediction to learn a compact but rich encoding of EEG signals. LaBraM likewise showed superior performance on varied tasks (abnormal EEG detection, event classification, emotion recognition, etc.), highlighting the benefits of cross-dataset pre-training. BIOT (Biosignal Transformer) further extended foundation modeling to multimodal biosignals~\citep{yang2023biot}, enabling cross-dataset learning by tokenizing each channel into a “sentence” of patch tokens; it demonstrated the feasibility of training one model on mixed EEG, ECG, and motion data. Despite these successes, a limitation of current EEG foundation models is their reliance on generic self-supervised objectives (e.g. masked autoencoding, sequence prediction) that are agnostic to any specific task. While these objectives ensure broad coverage of EEG variability, they might underemphasize fine-grained features crucial for particular tasks (for instance, the difference between EEG frequency bands is critical for motor imagery but a vanilla autoencoder may not prioritize that).

\section{Method}
\label{sec:Method}
\begin{figure}[t]
    \centering
    \includegraphics[width=\linewidth]{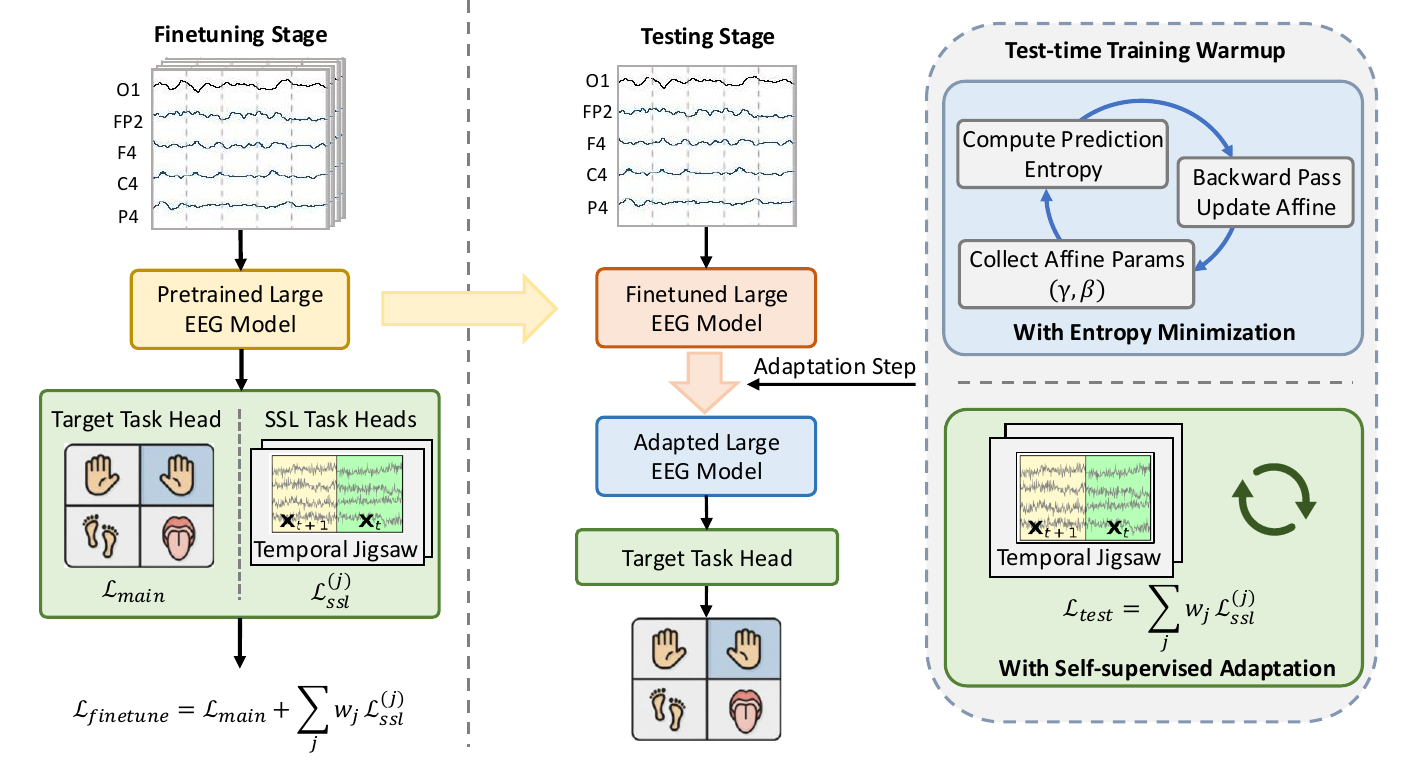}
    \caption{Overview of our NeuroTTT structure, illustrated on the Motor Imagery task. During fine-tuning, we augment the supervised head with two domain-specific self-supervised objectives: a Stopped-Band Prediction task plus a Temporal Jigsaw task. At inference, we perform self-supervised test-time training on individual samples or Tent-based entropy minimization to continually calibrate the model to each new input.}
    \label{fig:PPL}
\end{figure}

NeuroTTT comprises three interconnected components that together ensure a better alignment of the foundation model with each target EEG task: Stage I: Domain-Specific Self-Supervised fine-tuning, Stage II: Self-Supervised Test-Time Training, or Test-Time Entropy Minimization  at inference. An overview of the proposed alignment strategy is shown in Figure \ref{fig:PPL}. 

\subsection{Stage I — Domain-Specific Self-Supervised Fine-tuning}
\begin{spacing}{0.95}
In the preliminary stage, we adapt the pre-trained EEG foundation model to a specific downstream task by self-supervised fine-tuning. The pre-trained model (e.g. CBraMod or LaBraM) used as a shared feature extractor, feeding into different downstream branches: (a) the main task branch, which is a classifier for the labeled downstream task, and (b) two self-supervised task branches, each consisting of a small task-specific head dedicated to a particular unlabeled pretext task. All branches share the same foundation model backbone, but have separated objective functions. During fine-tuning, we jointly optimize the supervised loss on the main task and the losses of the self-supervised tasks. By doing so, we explicitly encourage the backbone’s representations to capture not only generic EEG features (from pre-training) but also task-relevant features that the self-supervised objectives focus on. This effectively aligns the feature space with the downstream task, preparing a better representation for the main-task.

Formally, we denote a pre-trained EEG foundation backbone $f_\theta$ with task head $g_\phi$. We jointly optimize a supervised loss and lightweight, EEG-aware SSL heads that share the backbone:

\begin{equation}
\mathcal{L}_{\text{finetune}} = \mathcal{L}_{\text{main}}\big(g_\phi(f_\theta(x)), y\big) 
+ \sum_j w_j \, \mathcal{L}_{\text{ssl}}^{(j)} \Big(h^{(j)}\big(f_\theta(\tilde{x}^{(j)})\big)\Big).
\end{equation}


where $w_j$ are weighting coefficients that balance the importance of each SSL task. $\tilde{x}^{(j)}$ is a self-supervised view that encodes EEG priors (e.g., \textbf{Stopped-Band Prediction}, \textbf{A--P Flip}, \textbf{Temporal Jigsaw}); heads $h^{(j)}$ are tiny classifiers.  In practice, we choose $w_j$ to ensure the self-supervised losses are on a comparable scale to the main loss; the additional tasks are intended to guide the representation without overwhelming the primary task objective.

We design domain-specific SSL tasks for each downstream application, informed by neuroscientific knowledge of task specific EEG signatures. One common self-supervision task across the three domains, is the Stopped Band Prediction task, with the frequency band divisions vary by downstream tasks. The second task is also specifically tailored to each domain. With an Amplitude Scaling Prediction for Imagined Speech Classification, an Anterior-Posterior Flip Detection for Mental Stress Detection and a Temporal Jigsaw Task for Motor Imagery.

\textbf{Mental Stress Detection – Anterior-Posterior Flip Detection:} Mental stress is known to induce characteristic EEG changes, often with strong asymmetry between the frontal and parietal regions of the scalp. Under stress or cognitive load, frontal lobe activity tends to increase relative to rear regions, engaging the frontoparietal network. To leverage this domain knowledge, we design an Anterior-Posterior (A–P) Flip self-supervised task. We define pairs of EEG channels that are roughly symmetric across the anterior-posterior axis of the head. We then randomly choose whether to swap these pairs (i.e. exchange frontal channel signals with their posterior counterparts) in a given EEG sample. The model’s SSL head must predict whether the input has been “flipped” or is in its original orientation (a binary classification). This pretext forces the model to learn the normal topographic distribution of activity for the task – specifically, to recognize the difference between a normal vs. front-back inverted EEG. By doing so, the encoder learns to encode the asymmetric topographic signature of stress-related brain activity and the coherence of the frontoparietal network without any explicit labels. This should yield a more transferable “stress representation” that is sensitive to spatial patterns known to correlate with mental workload or stress.

Details of the other SSL tasks can be found in Appendix \ref{sec:Appendix_ssl_tasks}. After adding these tasks, the fine-tuning process optimizes the combined loss $\mathcal{L}_{finetune}$ over the training set of the target task. The outcome is a model that still leverages the general knowledge from pre-training but is specialized with domain-specific feature awareness. Because the SSL tasks are designed to be lightweight (predicting simple transformations), they do not require many extra parameters (each SSL head is a small classification layer) and do not need ground-truth labels from the dataset. They can be seen as a form of feature guiding for the model during fine-tuning. Once training is complete, we can either keep these SSL heads for potential use at test time (as we describe next) or discard them if only the main task prediction is needed during inference.

\end{spacing}

\subsection{Stage II — Test-Time Training with Self-Supervision}
\begin{spacing}{0.95}

While the preliminary fine-tuning aligns the model to the general characteristics of the task domain, there can still be residual distribution shifts at test time. For example, differences in EEG noise level, electrode impedance, or individual brain anatomy can cause a drop in performance. To mitigate this, we incorporate Self-Supervised Test-Time Training (TTT) as an online adaptation mechanism~\citep{xiao2024beyond}.

The key idea of TTT is to leverage a domain-specific self-supervised objective on each unlabeled test sample to briefly adapt the model before making the final prediction. Given a new test input $\mathbf{x}_{test}$ with no ground-truth label, our TTT procedure is:

\begin{itemize}

    \item \textbf{Maintain Self-Supervised Head:} We retain the self-supervised tasks heads (learned during fine-tuning) for use during inference. This head produces a prediction for the self-supervised pretext task (e.g. predicting a rejected band or a Anterior-Posterior flipped indicator) on the test sample.
    
    \item \textbf{TTT Warm-Up (Adaptation Step):} When a new unlabeled test sample (or a small batch of samples) is received, we first perform a forward pass on $\mathbf{x}_{test}$ to compute the self-supervised loss $L_{SSL}(\mathbf{x}_{test}; \theta)$, since the main task loss cannot be computed without a label. For example, $L_{SSL}$ could be a cross-entropy loss for correctly predicting an augmented transformation of $\mathbf{x}_{test}$. We then update the model parameters by taking a single gradient descent step to minimize this loss, yielding adapted parameters $\theta'$:
    \begin{equation}
        \theta' = \theta - \alpha \nabla_{\theta} L_{SSL}(\mathbf{x}_{test}; \theta).
    \end{equation}
    where $\alpha$ is a small learning rate. This adaptation step personalizes the model to the test sample by tuning it to current sample’s idiosyncrasies (such as its noise characteristics or signal scale).
    
    \item \textbf{Inference after Adaptation:} After this brief adaptation on the unlabeled sample, we perform a forward pass of the adapted model on the same input (this time through the main task head) to get the final prediction (e.g. predicted class of imagined speech, stress level, or motor imagery label). We then reset the model to its original state before the next sample, or in an online scenario, we carry the updated state forward.
\end{itemize}

This procedure effectively personalizes the model to each test input on the fly. Intuitively, by solving a self-supervised puzzle on the test EEG, the model “tunes” itself to that EEG’s idiosyncrasies (noise distribution, amplitude scaling, etc.) in a way that also aligns with the main task features. Our approach builds on the method proposed by \cite{sun2020test} for image classification, but here applied to time-series brain data. We emphasize that no human labels are used in this adaptation, the model is optimizing a purely self-generated objective.
\end{spacing}

\subsection{Stage II — Test-Time Entropy Minimization}
\begin{spacing}{0.95}
Alongside self-supervised TTT, we incorporate Test-Time Entropy Minimization, often referred to as Tent adaptation. This is a complementary approach that does not rely on a separate SSL head, but instead uses the model’s main output entropy as a guidance for adaptation. The principle, introduced by \cite{wang2020tent}, is that for an unlabeled test example, a confident prediction (low entropy of the output softmax) is generally more likely to be correct, assuming the model’s learned decision boundaries are relatively well-aligned. Therefore, one can adapt the model on test data by directly minimizing the entropy of its predictions. 

Formally, for a given test sample $\mathbf{x}{test}$, we define the entropy of the model’s predictive distribution $p{\theta}(y \mid \mathbf{x}_{test})$ as:

\begin{equation}
    L_{ent}(\theta; \mathbf{x}_{test}) 
= - \sum_{c=1}^{C} p_{\theta}(y=c \mid \mathbf{x}_{test}) 
\log p_{\theta}(y=c \mid \mathbf{x}_{test}),
\end{equation}

where $C$ is the number of output classes. In our implementation, we apply Tent as follows: when a batch of test data is observed, we feed it through the model’s main task head to obtain predictions (probability distribution over classes). We compute the entropy of these predictions and take a gradient step to reduce $L_{ent}$ for adjusting the model parameters to minimize this entropy. Crucially, to preserve stability, we follow the established practice of updating only the Batch Normalization (BN) parameters during this process, while keeping the other weights fixed:

\begin{equation}
 \theta_{BN} \leftarrow \theta_{BN} - \alpha \nabla_{\theta_{BN}} L_{ent}(\theta; \mathbf{x}_{test}).
\end{equation}

Here $\theta_{BN}$ denotes the collection of affine scale and shift parameters (and, optionally, the running mean and variance) of all BatchNorm layers. By adjusting only these normalization parameters, the model recalibrates its internal feature statistics to better match the current test batch, thereby lowering its prediction uncertainty.

The Tent adaptation is extremely lightweight – it adds negligible computational cost (just one forward-backward pass per batch) and only modifies a tiny fraction of the model’s parameters. This makes it well-suited for real-time adaptation scenarios and ensures that the risk of overfitting is low (since BN layers have limited capacity compared to the entire network). We found that using entropy minimization at test time yields even better results than self-supervised TTT in some cases, especially for cross-subject transfer. This somewhat counter-intuitive outcome is likely because the full-model TTT (as above) adapts many parameters on a single sample, which, if the sample is noisy or unrepresentative, could momentarily destabilize the model. In contrast, Tent’s BN-only, entropy-driven update provides a gentler nudging of the model towards the target domain’s feature distribution, maintaining overall stability of the learned representations. Additionally, Tent naturally works in an online continual adaptation setting (updating BN for each batch) without needing to reset the model state between samples.

\end{spacing}

\section{Experiments}
We conduct experiments to evaluate the effectiveness of our proposed alignment strategies on multiple EEG datasets and foundation models. We first describe the experimental setup, including the pre-trained models used, the downstream tasks and datasets, and implementation details such as preprocessing. We then present quantitative results comparing various fine-tuning and adaptation methods, followed by ablation studies that dissect the contribution of each component of our approach.

\subsection{Experimental Setup}
\begin{spacing}{0.95}

\textbf{Pre-trained Foundation Models:} We evaluate our methods using two representative state-of-the-art large-scale EEG foundation models, each with distinct architectures and pre-training strategies, which provides a robust test of our alignment methods. Detailed descriptions of both models are provided in Appendix~\ref{sec:Appendix_models}. For all experiments, we use these foundation models as feature extractors and apply various adaptation strategies, as outlined in the subsequent sections. Both models internally downsample input signals to 200~Hz, thereby capturing frequency components up to 100~Hz in accordance with the Nyquist theorem.

\textbf{Downstream BCI Tasks and Datasets:} We evaluate on three diverse EEG decoding tasks to demonstrate generality: Imagined Speech Classification~\citep{jeong20222020}, Mental Stress Detection~\citep{goldberger2000physiobank,zyma2019electroencephalograms} and Motor Imagery Classification~\citep{brunner2008bci}. All the downstream EEG tasks with the corresponding datasets and preprocessing are presented in Appendix \ref{sec:Appendix_datasets}.

\textbf{Performance Metrics:} For the binary Mental Stress Detection task, we use Balanced Accuracy, Area Under the Precision-Recall Curve (AUC-PR), and Area Under the Receiver Operating Characteristic Curve (AUROC) as evaluation metrics, with AUROC designated as the monitoring score. For the multi-class tasks—Imagined Speech Classification and Motor Imagery Classification—we evaluate performance using Balanced Accuracy, Cohen’s Kappa, and Weighted F1 Score, with Cohen’s Kappa selected as the monitoring metric.

\textbf{Finetuning and Adaptation Strategies:} In all of our finetuning and adaptation settings, the foundation model remains fully trainable. This design choice is crucial for better aligning the pre-trained representations with the specific demands of each downstream EEG task. By allowing gradient updates throughout the model, we enable deeper integration of task-specific information, improving both accuracy and generalization. We compare the following strategies on each task:

\begin{itemize}
    \item \textbf{Full Supervised Finetuning:} Train the entire foundation model and a small task head (1--3 linear layers) end-to-end on labeled data. This is the de facto standard approach.
    \item \textbf{Lora Finetuning:} A parameter-efficient fine-tuning method that only introduces low-rank updates to a pre-trained model’s weights~ \citep{hu2022lora}, popular in LLMs finetuning. 
    \item \textbf{SHOT (Source Free Domain Adaptation):} A source-free domain adaptation setting for downstream tasks~\citep{liang2020we}. In essence, SHOT uses the source-trained classifier’s “hypothesis” and adjusts the feature extractor to better fit target data without source data. 
    \item \textbf{NeuroTTT:} Our full method. We fine-tune the model on the training set using our multi-task loss (with the appropriate domain-specific SSL tasks for each dataset) as described in the Method \ref{sec:Method} section. At inference, we apply test-time training with two variants: 

    (a) TTT with SSL, which performs a single gradient step on the SSL task for each test sample; (b) TTT with Tent, which applies entropy minimization to test batches by updating batch normalization statistics.
\end{itemize}

All models are optimized using the Adam optimizer~\citep{kingma2014adam}. For full-parameter finetuning—including any LoRA layers or SSL heads—we use a learning rate of 1e-4. We train each model until validation performance converges. For Mental Stress Detection and Motor Imagery Classification, convergence typically occurs within 20 epochs. In contrast, Imagined Speech Classification requires longer training due to slower convergence under domain-specific SSL supervision; we train for 100 epochs under supervised-only settings and extend to 150 epochs when incorporating domain-specific SSL. More hyperparameter setting can be found in Appendix \ref{sec:Appendix_Hyperparameters}.
\end{spacing}




\subsection{Results}

Table \ref{tab:results_BCIC2020} summarizes the results for Imagined Speech Classification, tested on BCI Competition 2020-III dataset. Table \ref{tab:results_ma} reports performance in Mental Stress Detection task, tested on MentalArithmetic Stress dataset. Table \ref{tab:results_bciciv2a} shows the cross-subject results in Motor Imagery tasks, tested on BCI Competition IV-2a dataset.

\begin{table}[htbp]
    \captionsetup{skip=2pt}
    \caption{Results on BCIC2020-III (5-class).}
    \centering
    \scriptsize
    \setlength{\tabcolsep}{4pt}
    \renewcommand{\arraystretch}{1.1}
    \resizebox{\textwidth}{!}{%
    \begin{tabular}{l|ccc|ccc} 
    \toprule
    \multirow{2}{*}{\makecell[l]{\textbf{Aligning}\\\textbf{Strategies}}} &
    \multicolumn{3}{c}{\textbf{CBraMod}~\citep{wang2024cbramod}} &
    \multicolumn{3}{c}{\textbf{LaBraM-Base}~\citep{jiang2024large}} \\
    \cmidrule(lr){2-4}\cmidrule(lr){5-7}
    & \textbf{Balanced-Accuracy} & \textbf{Cohen’s Kappa} & \textbf{Weighted F1}
    & \textbf{Balanced-Accuracy} & \textbf{Cohen’s Kappa} & \textbf{Weighted F1} \\
    \midrule
    Linear & 0.3976 $\pm$ 0.0113 & 0.2470 $\pm$ 0.0141 & 0.3968 $\pm$ 0.0120 & 0.2529 $\pm$ 0.0260 & 0.0661 $\pm$ 0.0325 & 0.2290 $\pm$ 0.0565 \\
    Shallow MLP & 0.4733 $\pm$ 0.0098 & 0.3417 $\pm$ 0.0122 & 0.4735 $\pm$ 0.0098 & 0.2493 $\pm$ 0.0315 & 0.0517 $\pm$ 0.0394 & 0.2304 $\pm$ 0.0442 \\
    Lora & 0.3531 $\pm$ 0.0158 & 0.1913 $\pm$ 0.0198 & 0.3525 $\pm$ 0.0159 & 0.2484 $\pm$ 0.0372 & 0.0606 $\pm$ 0.0215 & 0.2374 $\pm$ 0.0410 \\
    SHOT & 0.5008 $\pm$ 0.0010 & 0.3760 $\pm$ 0.0013 & 0.5004 $\pm$ 0.0011 & 0.2493 $\pm$ 0.0283 & 0.0717 $\pm$ 0.0229 & 0.2433 $\pm$ 0.0274 \\
    \midrule
    TTT with SSL & 0.5887 $\pm$ 0.0170 & 0.4858 $\pm$ 0.0212 & 0.5886 $\pm$ 0.0164 & 0.2720 $\pm$ 0.0083 & 0.0900 $\pm$ 0.0104 & 0.2649 $\pm$ 0.0137 \\
    TTT with Tent & \textbf{0.5898 $\pm$ 0.0148} & \textbf{0.4872 $\pm$ 0.0185} & \textbf{0.5895 $\pm$ 0.0142} & \textbf{0.2742 $\pm$ 0.0047} & \textbf{0.0928 $\pm$ 0.0159} & \textbf{0.2713 $\pm$ 0.0126} \\
    \bottomrule
    \end{tabular}
    }

    \label{tab:results_BCIC2020}
\end{table}

\begin{table}[ht]
\captionsetup{skip=2pt}
\caption{Results on MentalArithmetic.}
\centering
\scriptsize
\setlength{\tabcolsep}{4pt}
\renewcommand{\arraystretch}{1.1}
\resizebox{\textwidth}{!}{%
\begin{tabular}{l|ccc|ccc}
\toprule
\multirow{2}{*}{\makecell[l]{\textbf{Aligning}\\\textbf{Strategies}}} &
\multicolumn{3}{c}{\textbf{CBraMod}~\citep{wang2024cbramod}} &
\multicolumn{3}{c}{\textbf{LaBraM-Base}~\citep{jiang2024large}} \\
\cmidrule(lr){2-4}\cmidrule(lr){5-7}
& \textbf{Balanced-Accuracy} & \textbf{AUC-PR} & \textbf{AUROC}
& \textbf{Accuracy} & \textbf{AUC-PR} & \textbf{AUROC} \\
\midrule
Linear & 0.6389 $\pm$ 0.0092 & 0.4880 $\pm$ 0.0262 & 0.7788 $\pm$ 0.0408 & 0.6123 $\pm$ 0.0254 & 0.4826 $\pm$ 0.1124 & 0.6865 $\pm$ 0.0523 \\
Shallow MLP    & 0.6486 $\pm$ 0.0310  & 0.4459 $\pm$ 0.0527  & 0.7301 $\pm$ 0.0379  & 0.6125 $\pm$ 0.0249 & 0.2537 $\pm$ 0.0575 & 0.7329 $\pm$ 0.0191 \\
Lora   & 0.6001 $\pm$ 0.0144 & 0.5080 $\pm$ 0.0297 & 0.7327 $\pm$ 0.0239 & 0.5944 $\pm$ 0.0136 & 0.2475 $\pm$ 0.0245 & 0.7362 $\pm$ 0.0085 \\
SHOT & 0.6308 $\pm$ 0.0297 & 0.6063 $\pm$ 0.0613 & 0.7696 $\pm$ 0.0067 & 0.5827 $\pm$ 0.0274 & 0.4783 $\pm$ 0.0343 & 0.5826 $\pm$ 0.0267 \\
\midrule
TTT with SSL  & 0.7196 $\pm$ 0.0553 & 0.5720 $\pm$ 0.0798 & 0.8081 $\pm$ 0.0652 & 0.7161 $\pm$ 0.0231 & 0.6494 $\pm$ 0.0337 & 0.8180 $\pm$ 0.0081 \\
TTT with Tent & \textbf{0.7280 $\pm$ 0.0509}  & \textbf{0.5925 $\pm$ 0.0615}  & \textbf{0.8256 $\pm$ 0.0409}  & \textbf{0.7269 $\pm$ 0.0106} & \textbf{0.6401 $\pm$ 0.0402} & \textbf{0.7916 $\pm$ 0.0475} \\
\bottomrule
\end{tabular}
}

\label{tab:results_ma}
\end{table}

\begin{table}[!ht]
\captionsetup{skip=2pt}
\caption{Results on BCIC-IV-2a.}
\centering
\scriptsize
\setlength{\tabcolsep}{4pt}
\renewcommand{\arraystretch}{1.1}
\resizebox{\textwidth}{!}{%
\begin{tabular}{l|ccc|ccc}
\toprule
\multirow{2}{*}{\makecell[l]{\textbf{Aligning}\\\textbf{Strategies}}} &
\multicolumn{3}{c}{\textbf{CBraMod}~\citep{wang2024cbramod}} &
\multicolumn{3}{c}{\textbf{LaBraM-Base}~\citep{jiang2024large}} \\
\cmidrule(lr){2-4}\cmidrule(lr){5-7}
&\textbf{Balanced-Accuracy} & \textbf{Cohen’s Kappa} & \textbf{Weighted F1}
&\textbf{Balanced-Accuracy} & \textbf{Cohen’s Kappa} & \textbf{Weighted F1} \\
\midrule
Linear & 0.5028 $\pm$ 0.0230 & 0.3370 $\pm$ 0.0307 & 0.4809 $\pm$ 0.0293 & 0.41111 $\pm$ 0.00285 & 0.21482 $\pm$ 0.00382 & 0.34972 $\pm$ 0.03412 \\
Shallow MLP    & 0.4578 $\pm$ 0.0317 & 0.2771 $\pm$ 0.0423 & 0.4535 $\pm$ 0.0315 & 0.34479 $\pm$ 0.02853 & 0.12639 $\pm$ 0.03721 & 0.22937 $\pm$ 0.01984 \\
Lora   & 0.4164 $\pm$ 0.0090 & 0.2218 $\pm$ 0.0122 & 0.3952 $\pm$ 0.0227 & 0.34375 $\pm$ 0.03313 & 0.12500 $\pm$ 0.04472 & 0.27821 $\pm$ 0.04114 \\
SHOT & 0.5354 $\pm$ 0.0267 & 0.3766 $\pm$ 0.0365 & 0.5228 $\pm$ 0.0346 & 0.36310 $\pm$ 0.02880 & 0.36310 $\pm$ 0.02880 & 0.15080 $\pm$ 0.03840 \\
\midrule
TTT with SSL  & 0.5435 $\pm$ 0.0112 & 0.3914 $\pm$ 0.0149 & 0.5249 $\pm$ 0.0161 & 0.43835 $\pm$ 0.01856 & 0.25120 $\pm$ 0.01146 & 0.37380 $\pm$ 0.05552 \\
TTT with Tent & \textbf{0.5674 $\pm$ 0.0062} & \textbf{0.4232 $\pm$ 0.0082} & \textbf{0.5537 $\pm$ 0.0068} & \textbf{0.45110 $\pm$ 0.01950} & \textbf{0.26819 $\pm$ 0.01257} & \textbf{0.41559 $\pm$ 0.05358} \\
\bottomrule
\end{tabular}
}

\label{tab:results_bciciv2a}
\end{table}

As shown in the tables, our method pushes Large Brain Models to a markedly higher performance level.
Notably, LoRA shows the worse performance among all methods. This partial adaptation can struggle to align pre-trained EEG features to a new task because the frozen backbone’s representations remain largely unchanged. If the pre-trained features are not already well-suited for the target EEG task, a small low-rank tweak may be insufficient to bridge the gap. In EEG applications, tasks and datasets often differ substantially (e.g. different cognitive tasks, pathologies, or recording setups), so the foundation model’s features may not cleanly separate the new classes without significant reorientation. Since LoRA can only adjust the model via a limited subspace of parameters, it may fail to reshape the feature space enough for the downstream classification, leaving misalignments between what the backbone encodes and what the new task requires. Recent work on EEG foundation models has indeed observed that parameter-efficient “additive” tuning methods (like adapters or LoRA) yield inconsistent performance across tasks and pre-trained models. These methods often heavily depend on how closely the pre-training data matches the downstream domain. In other words, if the EEG foundation model wasn’t pre-trained on data very similar to the target task, a LoRA-based adaptation might not adequately realign the representation to the new task’s patterns. This is one reason LoRA underperforms: it cannot fully compensate when the pretrained feature space and the downstream data distribution are misaligned. Addressing this misalignment is one of the central problems our method tackles.

Unlike the large pretrain–downstream gap, the distribution shift between subjects is typically less severe. In an already well-learned feature space, applying full-parameter SSL fine-tuning can actually mislead the representation, especially when the incoming EEG is noisy or unrepresentative. By contrast, Tent’s BN-only, entropy-driven updates provide a gentler, more controlled push toward the target domain’s distribution, preserving the overall stability of the learned features. This explains why Tent often outperforms test-time SSL in our evaluations.

\subsection{Ablation Study:}
\begin{figure}
    \centering
    \includegraphics[width=0.9\linewidth]{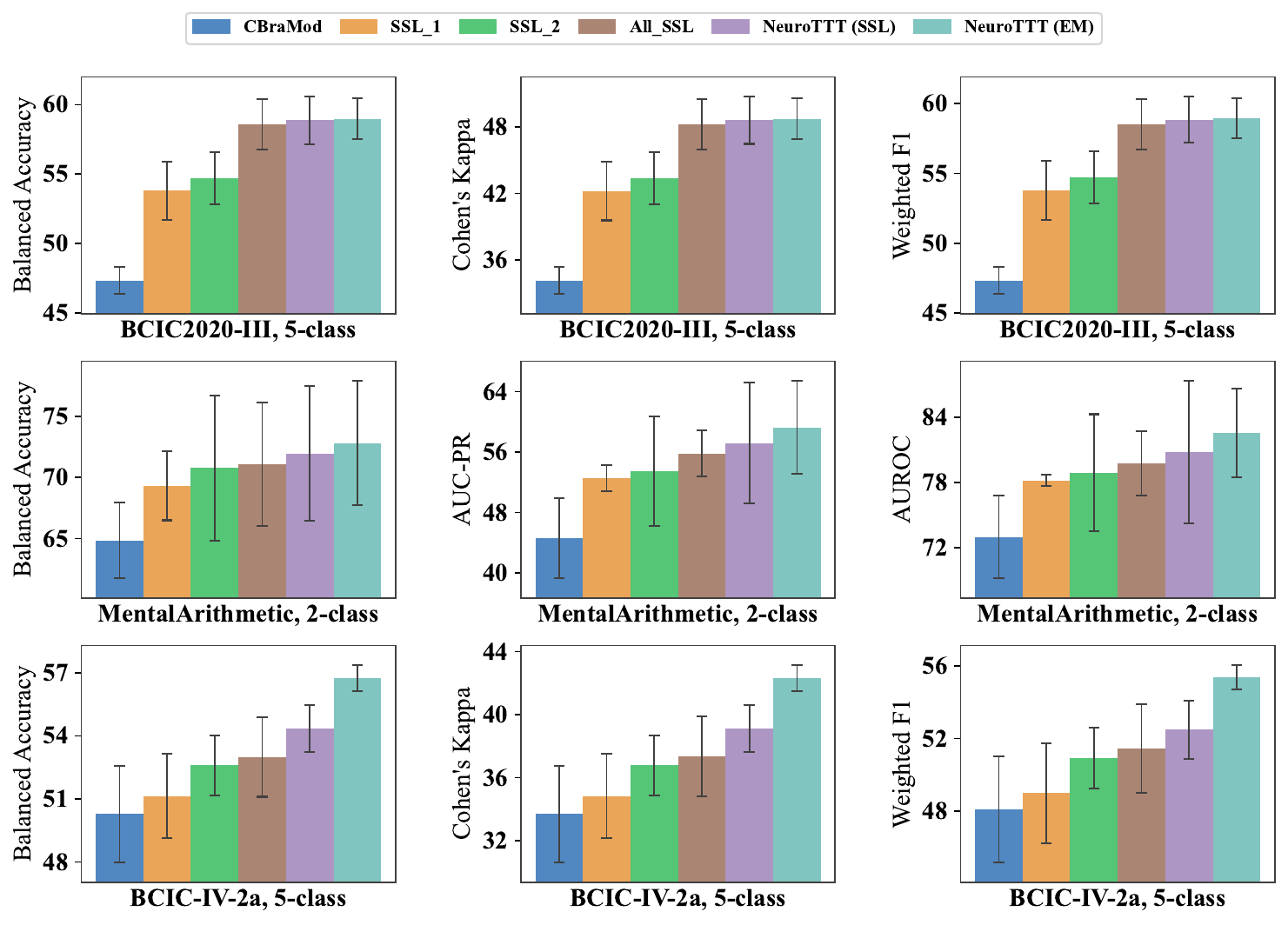}
    \caption{Ablation Study of NeuroTTT on CBraMod.}
    \label{fig:Ablation}
\end{figure}

\begin{spacing}{0.9}
We perform ablation studies to validate the contribution of each component of our approach: each domain-specific self-supervised finetuning tasks and the test-time training mechanisms, as shown in Figure \ref{fig:Ablation}. The ablation experiment is designed as follows:



\textbf{Isolated SSL Task Fine-tuning.} To assess the contribution of each self-supervised objective, we fine-tuned the model with each SSL task in isolation. The results show that every domain-specific SSL task injects distinct, informative signals into the representation; combining them produces additive, cumulative gains.

\textbf{Effect of Test-Time Training (TTT).} We compared models with vs. without TTT to evaluate its effectiveness. Notably, TTT delivers modest improvements on the BCIC dataset (within-subject, non–cross-subject), but yields substantial benefits on the two cross-subject tasks—exactly as expected given the stronger distribution shifts across subjects.
\end{spacing}



\section{Conclusion}
\begin{spacing}{0.9}
We proposed NeuroTTT: a comprehensive approach to aligning large EEG foundation models with domain-specific downstream tasks through a combination of targeted self-supervised finetuning and test-time training techniques. By injecting task-relevant inductive biases (frequency bands, amplitude scaling, topographic orientation, temporal order), we substantially improved the suitability of pre-trained representations for specific BCI tasks. Further, by leveraging test-time training and entropy minimization, our models can seamlessly adapt to new inputs on the fly, addressing one of the biggest challenges in EEG applications: distribution shift.

Our results on three diverse tasks and two state-of-the-art foundation models demonstrate that the proposed alignment strategy yields state-of-the-art performance, outperforming conventional fine-tuning and adaptation methods. The success of our approach suggests that bridging the gap between foundation model pre-training and target task requirements is both necessary and highly effective in the EEG domain. Rather than treating a foundation model as a monolithic feature extractor, we show that a small amount of task-specific “attention” in the form of self-supervision can unlock significant gains. Additionally, this work is a step toward more generalizable and personalized EEG AI systems: models that come with broad knowledge but can specialize themselves to each user or context safely and efficiently.

\end{spacing}


\bibliography{iclr2026_conference}
\bibliographystyle{iclr2026_conference}

\appendix

\vfill\pagebreak

\section{Domain-Specific Self-Supervised Fine-tuning}

\subsection{Domain-Specific Self-Supervised Tasks}
\label{sec:Appendix_ssl_tasks}
In addition to Anterior-Posterior Flip Detection, the other self-supervised tasks are as follows:

\textbf{Stopped Band Prediction: }The band-prediction task is motivated by the fact that EEG signals are composed of multiple frequency bands (delta, theta, alpha, beta, gamma, etc.), which often correlate with cognitive states. \cite{jayalath2024brain} employ band rejection during MEG pretraining and demonstrate that band prediction enables the model to learn frequency-aware representations more effectively. To leverage this, we apply a band-stop filter to the input EEG in a random frequency band (removing or attenuating one band’s power) and train the model to identify which band was removed. This encourages the model to learn representations that are sensitive to frequency-specific information, effectively tuning it to the spectral features crucial for specific downstream tasks. The band-prediction task is framed as a classification problem, where each class corresponds to the index of a removed frequency band. During fine-tuning, we randomly reject one band and compute a cross-entropy loss between the band-prediction head’s output and the true index of the dropped band. This task is lightweight and does not require any additional data beyond the unlabeled EEG sample itself.

\begin{table}[h]
\centering
\small
\captionsetup{skip=2pt}
\caption{Functional frequency bands in Imagined Speech Classification.}
\begin{tabular}{lll}
\toprule
\textbf{Band} & \textbf{Hz} & \textbf{Association} \\
\midrule
\multirow{2}{*}{Delta--Theta ($\delta/\theta$)} & \multirow{2}{*}{0.5--8}
 & Tracks the speech envelope and syllable-rate rhythms; aligns parsing \\
 &  & parsing at syllabic/prosodic timescales~\citep{giraud2012cortical}. \\
\addlinespace[2pt]
\multirow{2}{*}{Alpha--Beta ($\alpha/\beta$)} & \multirow{2}{*}{8--30}
 & $\alpha$: inhibitory gating of task-irrelevant regions; $\beta$: top-down \\
 &  & prediction and maintenance of internal models~\citep{jensen2010shaping}. \\
\addlinespace[2pt]
\multirow{2}{*}{Low Gamma ($\gamma_{\text{low}}$)} & \multirow{2}{*}{30--70}
 & Local cortical computations supporting feature integration during \\
 &  & speech analysis~\citep{giraud2012cortical}. \\
\addlinespace[2pt]
\multirow{2}{*}{High Gamma ($\gamma_{\text{high}}$)} & \multirow{2}{*}{70--100}
 & Auditory-cortex high-$\gamma$ encodes acoustic--phonetic detail predictive \\
 &  & of intelligibility~\citep{pasley2012reconstructing}. \\
\bottomrule
\end{tabular}
\label{tab:band_speech}
\end{table}

\begin{table}[ht]
\centering
\small
\captionsetup{skip=2pt}
\caption{Functional frequency bands in Mental Stress Detection.}
\begin{tabular}{lll}
\toprule
\textbf{Band} & \textbf{Hz} & \textbf{Association} \\
\midrule
\multirow{2}{*}{Frontal-midline Theta ($\theta$)} & \multirow{2}{*}{4--8}
 & Increases with working-memory load and cognitive-control \\
 &  & demands~\citep{jensen2002frontal}. \\
\addlinespace[2pt]
\multirow{2}{*}{Alpha ($\alpha$)} & \multirow{2}{*}{8--12}
 & Decreases with task engagement/alertness relative to relaxed \\
 &  &  rest~\citep{oken2006vigilance}. \\
\addlinespace[2pt]
\multirow{2}{*}{Low Beta ($\beta_{\text{low}}$)} & \multirow{2}{*}{13--20}
 & Indexes maintenance of the current sensorimotor/cognitive and \\ 
 &  & top-down control~\citep{engel2010beta}. \\
\addlinespace[2pt]
\multirow{2}{*}{High Beta ($\beta_{\text{high}}$)} & \multirow{2}{*}{20--30}
 & Strengthens with heightened vigilance/executive control; supports \\
 &  & long-range top-down communication~\citep{engel2010beta}. \\
\bottomrule
\end{tabular}
\label{tab:band_mental_stress}
\end{table}

\begin{table}[ht]
\centering
\small
\captionsetup{skip=2pt}
\caption{Functional frequency bands in Motor Imagery.}
\begin{tabular}{lll}
\toprule
\textbf{Band} & \textbf{Hz} & \textbf{Association} \\
\midrule
\multirow{2}{*}{Theta ($\theta$)} & \multirow{2}{*}{3--7}
 & Cognitive control/monitoring engaged during \\
 &  & MI preparation (frontal-midline $\theta$)~\citep{cavanagh2014frontal}. \\
\addlinespace[2pt]
\multirow{2}{*}{Mu ($\mu$, $\alpha$)} & \multirow{2}{*}{8--13}
 & Classic MI biomarker: contralateral $\mu$-ERD over sensorimotor cortex; \\
 &  & post-event $\mu$-ERS~\citep{pfurtscheller1999event}. \\
\addlinespace[2pt]
\multirow{2}{*}{Beta ($\beta$)} & \multirow{2}{*}{13--30}
 & $\beta$-ERD during MI/preparation; post-movement $\beta$ rebound (ERS) \\
 &  & reflects network reset~\citep{pfurtscheller1999event}. \\
\addlinespace[2pt]
\multirow{2}{*}{Low Gamma ($\gamma_{\text{low}}$)} & \multirow{2}{*}{30--45}
 & Brief motor-cortex $\gamma$ bursts reflecting fast local processing; \\
 &  & EEG analyses often cap at $\sim$45 Hz for SNR~\citep{cheyne2008self}. \\
\bottomrule
\end{tabular}
\label{tab:band_mi}
\end{table}

Stopped Band Prediction task is a shared self-supervision task across the three downstream tasks, with the frequency band divisions vary by tasks. The task-specific band divisions are summarized in Tables \ref{tab:band_speech}, \ref{tab:band_mental_stress}, \ref{tab:band_mi}. Both foundation models used in our study (CBraMod and LaBraM) were pre-trained at a sampling rate of 200 Hz. According to the Nyquist theorem, this limits the representable frequency range to approximately 100 Hz. Although higher frequency bands may be relevant for certain tasks like imagined speech decoding, our design is constrained by this 100 Hz upper limit.

\textbf{Imagined Speech Classification – Amplitude Scaling Prediction:} In imagined speech decoding (using the BCI Competition 2020-III dataset~\citep{jeong20222020}), an important feature is the amplitude modulation of EEG signals, as certain cognitive or speech imagery events can manifest as subtle amplitude changes in specific frequency ranges. We adopt an amplitude scaling pretext task, inspired by prior work on scaling transformations for speech detection~\cite{jayalath2024brain}. We artificially scale the raw EEG signal by a random factor $\alpha$ (drawn from a discrete set of scaling factors, e.g. 16 possible values from -2x to 2x) and task the model with predicting the scaling factor. This is implemented as a regression or classification problem (we treat it as classification over discrete scale indices). By learning to recognize how a scaled EEG differs from a normal one, the model becomes sensitive to the amplitude dynamics of the signal, which can improve its ability to detect the presence or absence of imagined speech patterns.  The amplitude of EEG signal is stretched or telescoped through 16 scale factors. where the scale-augmented signal can be expressed as $x_i^{st} = \alpha * x_i$. The objective is to predict the scaling factor $\alpha$.

\textbf{Motor Imagery – Temporal Jigsaw Task:} Motor imagery (using the BCI Competition IV-2a dataset~\citep{brunner2008bci}) entails a subject imagining limb movements, which produces EEG patterns with temporal dynamics (like event-related desynchronization in specific frequency bands following a cue). To ensure the model captures temporal order and dependencies, we introduce a temporal jigsaw task. We segment each EEG trial into a small number of consecutive time segments (for example, split into two or three chunks of equal length). We then randomly shuffle the order of these segments along the time axis and feed the jumbled sequence into the model. The model’s task is to predict the correct temporal order of the segments (essentially solving a jigsaw puzzle in time). In the simplest case of two segments, this reduces to a binary classification: whether the input is in correct chronological order or reversed. For more segments, we can have the model output a permutation or rank for each segment’s position. This task compels the model to learn temporal correlations and the progression of neural patterns over time – e.g. distinguishing the initial cue response period from the later motor imagery period. Mastering this pretext should enhance the model’s understanding of EEG time structure, benefiting tasks like motor imagery classification where timing (such as the onset of motor-related rhythms) is crucial.

\section{More Details for Experimental settings}

\subsection{Downstream BCI Tasks and Datasets}
\label{sec:Appendix_datasets}
Downstream BCI Tasks and Datasets: We evaluate on three diverse EEG decoding tasks:

\textbf{Imagined Speech Classification} – We use the BCI Competition 2020 Phase III (BCIC2020-3) dataset~\citep{jeong20222020}, which involves subjects imagining speaking specific phrases. The task is to classify which phrase is being imagined  (“hello”, “help me”, “stop”, “thank you” and “yes”). EEG signals were recorded at 64 channels and 256 Hz sampling rate. This is a within-subject classification problem: models are trained and tested on data from the same subject (with a standard train/test split per subject provided by the competition). We adhere to the original competition data splits for consistency with prior work. We attempted a cross-subject evaluation for this dataset (training on some subjects and testing on others) but found that both CBraMod and LaBraM performed at chance level in that scenario, likely due to the extreme variability and limited data per subject in imagined speech. Therefore, we report results in the within-subject setting, averaging performance across subjects.

\textbf{Mental Stress Detection} – We use public MentalArithmetic Stress dataset~\citep{goldberger2000physiobank,zyma2019electroencephalograms}, in which subjects perform arithmetic tasks under time pressure (stress condition) or relaxed conditions. EEG signals were recorded at 20 channels and 500 Hz sampling rate. The goal is to detect whether a given EEG trial corresponds to a high-stress condition or a low-stress condition (binary classification). We use a strict cross-subject protocol follow CBraMod: Subject 1 to 28 are set as training set, subject 29 to 32 are validation set and subiect 33 to 36 are set to test set. Evaluating is demonstrated on completely held-out subjects, to test generalization to unseen individuals’ stress responses.

\textbf{Motor Imagery Classification} – We use the well-known BCI Competition IV 2a dataset~\citep{brunner2008bci}, which contains EEG recordings from subjects performing motor imagery of four classes (left hand, right hand, both feet, tongue). EEG signals were recorded at 22 channels and 250 Hz sampling rate. We follow the standard cross-subject evaluation: Subject 1-5, 6-7, 8-9 are used for training, validation, and test, respectively, the predefined splits from the competition.

These tasks cover a range of EEG signal characteristics (from cognitive/internal speech processes to emotional stress responses to sensorimotor rhythms) and vary in number of classes and difficulty. They provide a solid testbed for our alignment approach. 

\subsection{Preprocessing}
\label{sec:Appendix_preprocessing}
We follow the preprocessing pipelines recommended by the foundation model authors to ensure compatibility with the models. For CBraMod and LaBraM, EEG signals are re-referenced and band-pass filtered (follow the setting in CBraMod) and then downsampled to 200hz follow the pre-trained sampling rate. Then the signals will be segmented into short windows (1-second segments). We do not apply any task-specific feature engineering – the raw (preprocessed) signals are fed into the models. The same preprocessing is applied consistently across all adaptation methods for fairness.

\subsection{Foundation Models}
\label{sec:Appendix_models}

We test our methods with two recent large-scale EEG foundation models as backbones:

\textbf{CBraMod~\citep{wang2024cbramod}:} a Criss-Cross Brain Transformer model (Wang et al., 2025) that was pre-trained on 12 public EEG datasets using a masked patch reconstruction objective. CBraMod features a dual-stream attention architecture separating spatial and temporal attention, and it produces a 768-dimensional feature vector per EEG trial (after temporal pooling). We use the official pre-trained weights released by the authors.

\textbf{LaBraM~\citep{jiang2024large}:} the Large Brain Model which was pre-trained on over 2,500 hours of EEG from around 20 datasets. LaBraM employs a ViT-like backbone with an EEG-specific tokenizer (based on vector-quantized neural spectral codes) and was trained to predict masked neural code tokens. We use the official pre-trained weights released by the authors.

These two models are representative of current state-of-the-art EEG foundation models; they have complementary architectures and pre-training approaches, which provides a robust test of our alignment methods. For all experiments, we treat the foundation model as a fixed feature extractor initially and then apply the different adaptation strategies on top (as described below). 

\subsection{Implementation Details}
\label{sec:Appendix_Hyperparameters}

\textbf{Finetuning and Adaptation Strategies:} 

\begin{itemize}
    \item \textbf{Full Supervised Finetuning:} The entire foundation model, along with the task-specific classification head, is trained end-to-end using labeled downstream data with 1-3 linear layers. This serves as a strong baseline for comparison.
    \item \textbf{Lora Finetuning:} A parameter-efficient fine-tuning method that only introduces low-rank updates to a pre-trained model’s weights. We insert trainable rank-$r$ matrices in each attention projection (queries, keys, values) and in the feed-forward layers of each transformer block, following \cite{hu2022lora}. We allow these LoRA layers to be trained on the labeled data while the original weights are frozen. This efficiently adapts the backbone without full fine-tuning. We tried to apply LoRA to as many layers as possible (all self-attention and intermediate dense layers in CBraMod/LaBraM) to maximize adaptability.
    \item \textbf{SHOT (Source Free Domain Adaptation):} We simulate a source-free domain adaptation setting for downstream tasks. We first train a classifier on the source training set (treating the foundation model as fixed feature extractor, akin to linear probing on source). Then, at test time, we adapt this classifier using the SHOT method: we freeze the classifier’s weights and instead update the feature extractor (foundation model) on the target (unlabeled test) data by maximizing the mutual information of predictions and minimizing entropy, as described by \cite{liang2020we}. In essence, SHOT uses the source-trained classifier’s “hypothesis” and adjusts the feature extractor to better fit target data without source data. Notably, while source-free domain adaptation methods like SHOT have been actively explored in other biosignals (e.g., EMG), they remain underexplored for EEG. Only very recent efforts have applied SHOT-style techniques to shallow neural networks in EEG~\citep{wang2024lightweight,li2024spdim}, and, to our knowledge, there is virtually no work integrating SHOT within EEG finetuning.
    \item \textbf{Domain-SSL + Test-Time Training (TTT):} This is our full method. We fine-tune the model on the training set using our multi-task loss (with the appropriate domain-specific SSL tasks for that dataset) as described in the Method \ref{sec:Method} section. At inference, we apply test-time adaptation. In practice, we evaluate two variants: 

    (a) TTT with SSL, which performs a single gradient step on the SSL task for each test sample; 

    (b) TTT with Tent, which applies entropy minimization to test batches by updating batch normalization statistics.
\end{itemize}

For test-time adaptation, NeuroTTT (SSL) uses one gradient update per test sample on the self-supervised task, without online training. For Tent, we apply several updates per test batch to the batch normalization layers using momentum-based exponential moving averages. More hyperparameters for NeuroTTT are shown in Table \ref{tab:neurottt-hparams}.

\begin{table}[t]
\centering
\caption{Test Stage hyperparameters for NeuroTTT (SSL)}
\label{tab:neurottt-hparams}
\begin{tabular}{l l}
\toprule
\textbf{Hyperparameters} & \textbf{Settings} \\
\midrule
Steps & 1 \\
Batch size & 1 \\
Learning rate & $1\times10^{-5}$ \\
Dropout & 0.1 \\
Optimizer & Adam \\
online & False \\
\midrule
$w_1$ (Imagined Speech) & 0.6 \\
$w_2$ (Imagined Speech) & 0.6 \\
$w_1$ (Mental Stress) & 0.2 \\
$w_2$ (Mental Stress) & 0.1 \\
$w_1$ (Motor Imagery) & 0.1 \\
$w_2$ (Motor Imagery) & 0.8 \\
\bottomrule
\end{tabular}
\end{table}

\section{Limitations and Future Works}
\label{sec:future}

\textbf{Limitations:} 
While our results demonstrate consistent gains across tasks and backbones, several limitations merit discussion:
\begin{enumerate}
    \item Task-Engineered SSL Dependence. Our alignment relies on manually designed domain-specific SSL tasks (e.g., band prediction, A–P flip, temporal jigsaw). These choices encode neuroscientific priors but may not be universally optimal; subpar task design can bias representations or yield negligible gains. Automating SSL design or meta-selecting pretext tasks remains future work.
    \item Scope of TTT Parameter Updates. Our current TTT with SSL uses full-parameter updates during inference. While effective, this may be unnecessary or induce instability on noisy trials. Future work will evaluate restricted adaptation—e.g., updating only late blocks, normalization layers, or lightweight adapters—to better trade off stability, compute overhead, and accuracy.
    \item TTT Stability vs. Benefit Trade-off. Test-time self-supervised training can be sensitive to noise and atypical trials, occasionally perturbing well-formed features. Although Tent offers a more stable BN-only alternative, its adaptation capacity is limited to normalization statistics and may under-correct severe shifts.
\end{enumerate}

\textbf{Future Work:} We believe this framework can be extended to other neural recording modalities (MEG, sEEG, fNIRS) and other tasks not explored in this paper. Future work will explore automated ways of generating domain-specific SSL tasks (perhaps via meta-learning) and more advanced test-time adaptation schemes (e.g. continuous adaptation in lifelong settings). We also envision incorporating subject metadata or physiological priors into the adaptation to further enhance personalization. Ultimately, aligning brain foundation models to downstream tasks and users is crucial for making these models truly practical, and our work takes an important step in that direction.

\end{document}